\documentclass{article}
\usepackage[final]{neurips_2019}
\usepackage[utf8]{inputenc} 
\usepackage[T1]{fontenc}    
\usepackage{hyperref}       
\usepackage{url}            
\usepackage{booktabs}       
\usepackage{amsfonts}       
\usepackage{nicefrac}       
\usepackage{microtype}      
\usepackage{graphicx}
\usepackage{float}
\usepackage[FIGTOPCAP]{subfigure}

\title{Bayesian Recurrent Framework for Missing Data Imputation and Prediction with Clinical Time Series}
\author{Yang Guo, Zhengyuan Liu, Pavitra Krishnaswamy$^*$, Savitha Ramasamy\thanks{denotes equal contribution. 
\newline The authors would like to acknowledge grant funding for Digital Health from the Science and Engineering Research Council, A*STAR, Singapore (Project No. A1818g0044).} \\
  Institute for Infocomm Research\\
  Agency for Science, Technology and Research (A*STAR) Singapore\\
  \texttt{\{pavitrak,ramasamysa\}@i2r.a-star.edu.sg}} 

\begin{document}

\maketitle

\begin{abstract}
Real-world clinical time series data sets exhibit a high prevalence of missing values. Hence, there is an increasing interest in missing data imputation. Traditional statistical approaches impose constraints on the data-generating process and decouple imputation from prediction. Recent works propose recurrent neural network based approaches for missing data imputation and prediction with time series data. However, they generate deterministic outputs and neglect the inherent uncertainty. In this work, we introduce a unified Bayesian recurrent framework for simultaneous imputation and prediction on time series data sets. We evaluate our approach on two real-world mortality prediction tasks using the MIMIC-III and PhysioNet benchmark datasets. We demonstrate strong performance gains over state-of-the-art (SOTA) methods, and provide strategies to use the resulting probability distributions to better assess reliability of the imputations and predictions. 
\end{abstract}

\vspace{-3.5ex}

\section{Introduction}
Real-world healthcare datasets across acute and community care settings exhibit a high prevalence of missing values. Due to this challenge, predictive analysis with these data is often challenging, and insights mined from these data may not be most reliable \cite{liptonmodeling,5,1,3,10}. In particular, this problem is especially cumbersome for clinical time series data comprising longitudinal records of patient state. This has spurred a longstanding interest in missing data imputation methods \cite{5,BayesianMissing_2003,review_2010}. 

Traditional approaches for such problems have relied on statistical models and associated Bayesian inference paradigms \cite{multiple_2011,9,hastie2005elements}, but these require strong constraints on the data-generating process, and treat the imputation and prediction as independent tasks \cite{3,multiple_2011,EHRD_2013}. To overcome these limitations, recent works have proposed deep learning approaches using recurrent neural networks \cite{5,ML4H,Lipton2016,9,1}. While these methods learn directly from the data without imposing specific assumptions on the underlying processes, and show promise for accurate imputation and prediction on clinical time series, they provide deterministic outputs and neglect the uncertainty inherent to the task.

In this work, we present a unified Bayesian framework for imputation and prediction with multivariate clinical time series. We embed a Bayesian Recurrent Neural Network and a Bayesian Neural Network within a recurrent dynamical system for integrative missing value imputation and prediction. We characterize performance on mortality prediction tasks with the publicly available  MIMIC-III \cite{6} and PhysioNet \cite{7} benchmark data sets, and demonstrate strong performance improvements. We further show strong correlations between the variability and accuracy of the imputations. These results suggest that our approach adapts the imputation model to the uncertainty inherent in the modelling task, and offers a principled way to assess reliability of imputations and predictions.

\section{Methods}
\textbf{Problem Formulation:} We consider dataset $D$ of $N$ samples. Each sample is denoted as an input-output pair $(\mathbf{X}, Y)$, where $\mathbf{X} = \left({\mathbf{x}_{1}, \mathbf{x}_{2}, \ldots, \mathbf{x}_{M}}\right)$ denotes a multivariate time series input with $M$ features and $Y$ denotes its output. Each feature is a sequence of observations over $T$ time steps $\mathbf{x}_{i} = \left[{x^1_{i}, x^2_{i}, \ldots, x^T_{i}}\right]$. In practice, $\mathbf{X}$ may have missing values. A masking matrix $\mathbf{V}_{N \times M}$ represents the presence of missing values: $V_{i,j} = 0$ when $X_{i,j}$ is missing, otherwise $V_{i,j} = 1$. The objectives are: (a) impute the missing values in $\mathbf{X}$ and (b) predict $Y$ given $\mathbf{X}$.

\begin{figure}[htp]
\centering
{\includegraphics[width=\textwidth, height=0.45\textwidth]{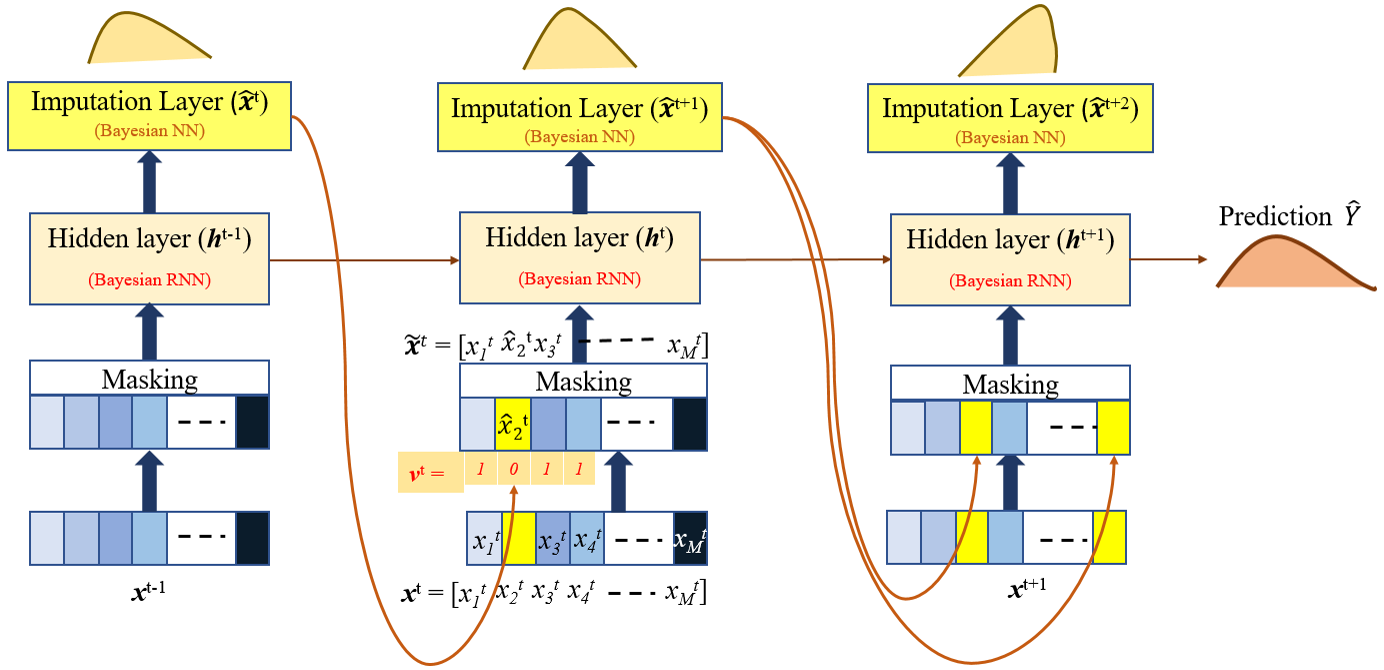}}
\caption{ \label{structure} Our Bayesian imputation framework contains 3 layers: masking, Bayesian RNN for prediction and Bayesian NN for imputation. Yellow blocks denote missing value. The arrows show the dynamical flow to estimate imputations $\mathbf{\tilde{X}}$, predictions $\hat{Y}$, and their associated distributions.}
\end{figure}

\textbf{Bayesian Recurrent Framework:} To address the above objectives, prior works leverage recurrent dynamical systems for deterministic models \cite{5, Lipton2016}. Here, we augment such recurrent dynamical systems with Bayesian approaches to model the uncertainty in the imputation and prediction tasks. Fig. \ref{structure} illustrates our proposed Bayesian recurrent framework for imputation and prediction. 

At each time step, the input is fed through the masking layer to a Bayesian recurrent neural network. The RNN hidden state dynamics are specified as:
\begin{equation}
\mathbf{h}^{t} = f(\mathbf{W}_{r}, \mathbf{h}^{t-1}, \mathbf{x}^t),
\end{equation}
where $\mathbf{h}^{t}$ is the RNN hidden state at time $t$ and $\mathbf{W}_{r}$ denotes the weights for the recurrent layer. As the inputs may not always be regularly sampled over time, we incorporate a temporal decay factor in the hidden state dynamics \cite{5}. To model the uncertainty in $f$, the Bayesian RNN considers a probabilistic distribution $p(\mathbf{W}_{r}|D)$ for the weights, instead of fixed values $\mathbf{W}_{r}$ \cite{8}. 

If $\mathbf{x}^t$ has missing values for any feature, we impute the missing values and replace $\mathbf{x}^t$ with:
\begin{equation}
\mathbf{\hat{x}}^{t} = g(\mathbf{U}_{x}, \mathbf{h}^{t-1}, \mathbf{x}^t),
\end{equation}
The imputation function $g$ implicitly models the correlation amongst features \cite{1}. To model the uncertainty in $g$, we use a Bayesian multilayer perceptron that considers a probabilistic distribution $p(\mathbf{U}_{x}|D)$ for the weights $\mathbf{U}_{x}$ \cite{8}.

We apply the imputations $\mathbf{\hat{X}}$ when the masking matrix $\mathbf{V}$ indicates missing values. Thus, across time steps, the updated input $\mathbf{\tilde{X}}$ is:
\begin{equation}
\mathbf{\tilde{X}} = (\mathbf{I}-\mathbf{V})\odot\mathbf{\hat{X}} + \mathbf{V}\odot\mathbf{X}
\end{equation}

With $\mathbf{\tilde{X}}$ and the RNN hidden states $\mathbf{H} = [\mathbf{h}^1, \mathbf{h}^2, \ldots, \mathbf{h}^T]$, the predicted output $\hat{Y}$ is:
\begin{equation}
\hat{Y}  = f_{\rm{out}}(\mathbf{\tilde{X}}, \mathbf{H})
\end{equation}
We use a linear form for $f_{out}$. The above equations specify the overall recurrent dynamical system. Then, to obtain the imputations and predictions given input $\mathbf{X}$, we need to compute the probabilistic distribution of the overall weights $\mathbf{W} = \left[\mathbf{W}_{r}, \mathbf{U}_{x}\right]$, given data $D$. As the true posterior distribution $p(\mathbf{W}|D)$ is intractable in general, we have to approximate it using Bayes by Backprop \cite{8}. Conceptually, Bayes by Backprop minimizes the Kullback-Leibler divergence between the approximate distribution $q(\mathbf{W}|\mathbf{\Theta})$ and the true posterior $p(\mathbf{W}|D)$. As such, the loss function for our estimation not only comprises the imputation and prediction errors, but also the KL-divergence loss, as below:
\begin{eqnarray}
L_{\rm{total}} &=& L_{\rm{KL}}(Y,\mathbf{X}, p(\mathbf{W})) + L_{\rm{imput}}(\mathbf{X}, \mathbf{\hat{X}}) + L_{\rm{pred}}(Y, \hat{Y}),\\
L_{\rm{KL}} &=& -\mathbb{E}_{q(\mathbf{W})}[\log p(Y|\mathbf{W}, \mathbf{X})] + KL[q(\mathbf{W}|\mathbf{\Theta}) || p(\mathbf{W})],\\
L_{\rm{imput}} &=&  (\mathbf{X} - \mathbf{\hat{X}})\odot\mathbf{V}, \\
L_{\rm{pred}} &=& L_{\rm{cross-entropy}}(\hat{Y},Y),
\end{eqnarray}
where $p(\mathbf{W})$ is the prior distribution for the weights, set as a mixed Gaussian. We highlight that the imputation loss only considers performance for sample values that are \emph{not} missing. The loss function controls the imputation, prediction and posterior distribution of the weights simultaneously. 

Minimizing the loss function, we obtain the posterior of the weights as well as the imputed values and output predictions. Our proposed framework adapts jointly to the uncertainty in the imputation and prediction process, functions as a regularizer to improve robustness, and also provides distributions of the resulting estimates for further study. 

\section{Experiments and Results}

\textbf{Data:} To evaluate our approach, we perform experiments on mortality prediction tasks using benchmark data sets from the PhysioNet 2012 Challenge  \cite{7} and the MIMIC-III collection \cite{6,3}. These data sets comprise of multivariate time series clinical features recorded from patients in the intensive care unit (ICU). For PhysioNet, the input comprises 35 numerical features in time series samples from 4000 admissions. For MIMIC-III, the input comprises 12 numerical features in time series samples from 14,681 admissions. In both cases, each input sample comprises 48 hourly time steps and the output is in-hospital mortality. We note that both data sets are sparse, with 78\% and 48\% of the values missing for PhysioNet and MIMIC-III, respectively. 

\textbf{Performance Metrics}: To enable evaluation of imputation performance, we simulate missingness at random (MAR) by eliminating 10\% of the known input values in each data sample\cite{1}. For this subset of data with simulated MAR, we evaluate imputation performance by computing the mean absolute error (MAE) and mean relative error (MRE). Further, we pick a test set with random subset of 20\% from all data samples to evaluate the prediction performance by computing the areas under the receiver operating characteristics and the precision recall curves (AUROC, AUPRC).

\textbf{Baselines:} We compare our method against three SOTA RNN based imputation methods: (a) Gated Recurrent Units (GRU-D) \cite{5}; (b) Recurrent Imputation in Time Series (RITS) \cite{1}; (c) Bayesian Recurrent Neural Networks (BRNN) that refers to a Bayesian RNN \cite{2} with all missing values imputed with zero. Unlike our method wherein the temporal decay factor only affects hidden states, the GRU-D baseline considers the decay factors both for input and hidden state dynamics. The RITS baseline considers a vanilla RNN without a Bayesian framework. Finally, the BRNN baseline helps evaluate the impact of a Bayesian approach for prediction, independently of data-driven imputation.  

\textbf{Performance Results:} Table \ref{Imputation_Performance} provides the performance results. Where previously reported results exist, we include them with citations. We also repeat the experiments for fair comparison, as required. We observe that our method outperforms the state-of-the-art techniques, beating the closest SOTA method by upto 2\% in imputation MRE and upto 3.3\% in prediction AUPRC. Our imputation performance improvement is less prominent in the MIMIC-III dataset, possibly as this dataset is less sparse than PhysioNet. We also characterized the imputation performance with increasing prevalence of simulated MAR for PhysioNet. With increasing MAR rates, our method offered increasing performance improvement over the closest SOTA method (RITS). Even with a 15\% increase in simulated MAR, our approach outperformed RITS by 3.5\% MRE. 

\begin{figure}
\centering
\subfigure[Distribution of imputations (one sample value)]{\includegraphics[width=0.32\textwidth]{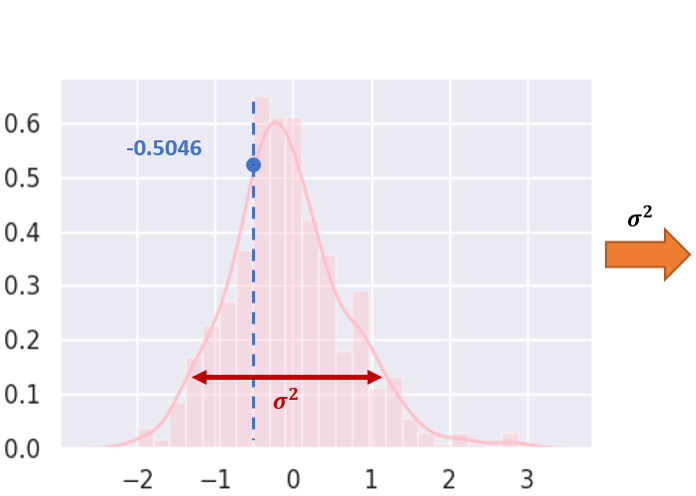}\label{distribution}}
\subfigure[Variability of imputations vs. MAE (across samples)]{\includegraphics[width=0.33\textwidth]{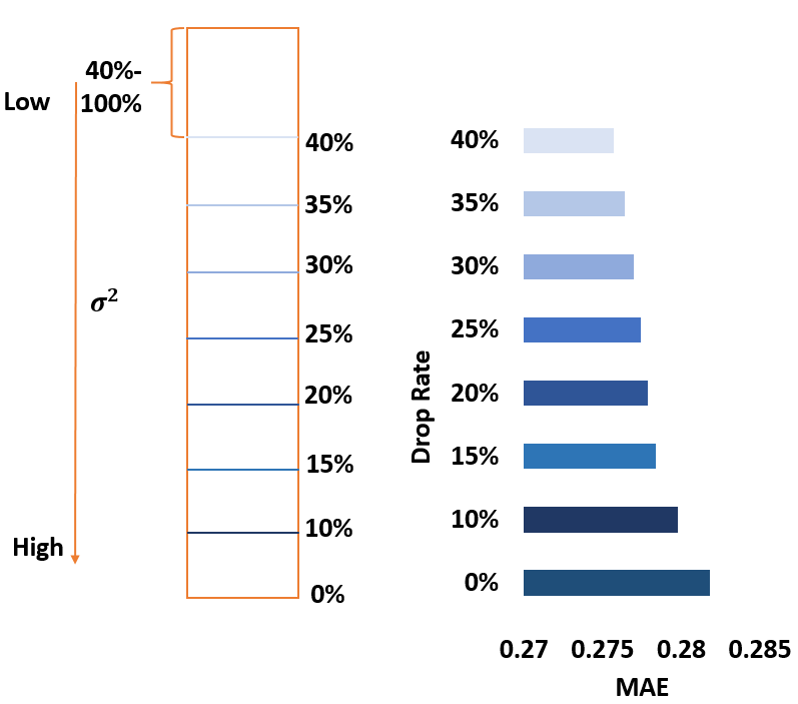}
\label{sample}}
\subfigure[Variability of imputations vs. MAE (by features)]{\includegraphics[width=0.33\textwidth]{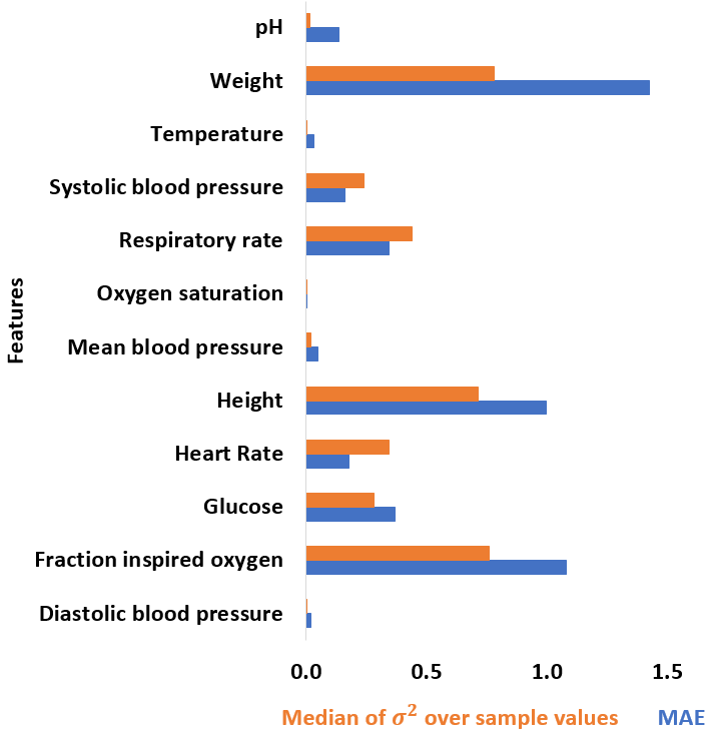}\label{feature}}
\caption{(a) Distribution of imputed values obtained by Monte Carlo iterations in our Bayesian framework. The variance $\sigma^2$ of the imputations is tightly correlated with MAE at sample level (b) and at feature level (c).}\label{figuncertainty}
\end{figure}

\textbf{Impact of Distributions:} One advantage of our Bayesian framework is that it provides the ability to get a distribution of imputations and predictions. First, for each simulated sample value, we used Monte Carlo iterations to obtain a distribution of the imputed values (Fig. \ref{distribution}). By visualizing the ground truth value atop the distribution (blue point), we can assess how far removed the distribution is from the ground truth. Second, we studied the variability implicit in the imputation process. For each of the simulated MAR, we obtain the variance of the distribution of imputed values, denoted by $\sigma^2$ (e.g., red line). We then sort the variances of all imputations in ascending order (Fig. \ref{sample}), eliminate missing values with the highest variances, and assess the impact on imputation MAE. We note, for example, that removing missing values with $\sigma^2$ values in the top 40 percentile of variances leads to lower MAE than when all missing values are considered. This suggests that there is a monotonic relation between the accuracy and variability of imputation. Third, we breakdown this accuracy vs variability relation to the individual feature level (Fig. \ref{feature}). The same trend carries through, wherein features with higher variability in imputed values (e.g., weight, height and inspired oxygen) tend to have higher MAE. These results suggest that, in real-world scenarios, when there is no ground truth, the variance $\sigma^2$ can serve as a means to assess reliability of the imputed values. 

\begin{table}[H]
\centering
\caption{\label{Imputation_Performance} Performance Comparison for Imputations and Predictions}
\renewcommand{\arraystretch}{1.05}
\begin{tabular}{lp{2.5cm}p{1.5cm}p{1.5cm}p{2.0cm}p{2.0cm}}
\hline
\toprule
\textbf{Dataset}     &\textbf{Methods}         & \textbf{MAE} & \textbf{MRE} & \textbf{AUROC} & \textbf{AUPRC}\\ \hline
PhysioNet &GRU-D \cite{1}           & $0.559$      & $0.776$      & $0.828$        & $-$            \\   \cline{2-6}
                     &RITS \cite{1}            & $0.300$      & $0.419$      & $0.840$      & $-$\\  \cline{2-6}
                     &BRNN                     & $0.708$      & $1.000$      & $0.811$        &$0.515$ \\  \cline{2-6}
                     &RITS                     & $0.296$      & $0.418$      & $0.834$        &$0.532$ \\  \cline{2-6}
                     &Ours                     & $\textbf{0.282}$ & $\textbf{0.398}$  & $\textbf{0.866}$  &$\textbf{0.553}$\\ \hline
MIMIC-III &GRU-D                    & $0.390$      & $0.779$      & $0.790$        &$0.421$  \\ \cline{2-6}
                     &BRNN                     & $0.503$      & $1.000$      & $0.775$        &$0.422$  \\ \cline{2-6}
                     &RITS                     & $0.151$      & $0.300$      &$0.805$        &$0.432$ \\ \cline{2-6}
                     &Ours                     & $\textbf{0.148}$    & $\textbf{0.294}$ & $\textbf{0.815}$    & $\textbf{0.465}$ \\ \hline
\bottomrule
\end{tabular}
\end{table}

\section{Discussion and Future Work}
We have developed a Bayesian recurrent framework to enable missing data imputation and prediction on clinical time series data sets. Our approach improves both imputation and prediction performance and is robust to increasing MAR. Further, by providing explicit probability distributions of imputed values and output predictions, it enables assessment of variability and reliability of the imputed values. This has important implications in real-world scenarios, where ground truth is lacking. Future work will consider expansions to include categorical features, varying length time series, different missingness patterns such as NMAR (Missing Not at Random) and MCAR (Missing Completely at Random), and develop more rigorous theoretical grounding. 

\bibliographystyle{plain}
\bibliography{ref.bib}

\end{document}